\documentclass[conference]{IEEEtran}
\IEEEoverridecommandlockouts

\usepackage{amsmath,amsfonts}
\usepackage{array}
\usepackage{textcomp}
\usepackage{stfloats}
\usepackage{url}
\usepackage{hyperref}
\usepackage{verbatim}
\usepackage{graphicx}
\usepackage{xcolor}
\usepackage{multirow}
\usepackage{graphicx,xfp}
\usepackage{booktabs,threeparttable}
\usepackage{tikz}
\usepackage{tikzscale}
\usepackage{comment}

\usepackage{caption}
\usepackage{subcaption}

\usepackage{pgfplots}
\usepgfplotslibrary{groupplots}
\usetikzlibrary{positioning}
\usetikzlibrary{external}
\usepackage{algorithm}
\usepackage[noend]{algpseudocode}

\newtheorem{remark}{Remark}

\newtheorem{definition}{Definition}[section]

\newcommand{\ie}{\emph{i.e.}}

\begin{document}

\title{Evaluating COVID-19 Sequence Data Using Nearest-Neighbors Based Network Model}

\author{\IEEEauthorblockN{Sarwan Ali}
\IEEEauthorblockA{
Department of Computer Science \\ Georgia State University, Atlanta, USA \\
sali85@student.gsu.edu} 
\thanks{2022 IEEE International Conference on Big Data (Big Data) | 978-1-6654-8045-1/22/\$31.00 \copyright 2022 European Union}
}

\maketitle

\begin{abstract}
The SARS-CoV-2 coronavirus is the cause of the COVID-19 disease in humans. Like many coronaviruses, it can adapt to different hosts and evolve into different lineages. It is well-known that the major SARS-CoV-2 lineages are characterized by mutations that happen predominantly in the spike protein. Understanding the spike protein structure and how it can be perturbed is vital for understanding and determining if a lineage is of concern. These are crucial to identifying and controlling current outbreaks and preventing future pandemics. Machine learning (ML) methods are a viable solution to this effort, given the volume of available sequencing data, much of which is unaligned or even unassembled. However, such ML methods require fixed-length numerical feature vectors in Euclidean space to be applicable. Similarly, euclidean space is not considered the best choice when working with the classification and clustering tasks for biological sequences.
For this purpose, we design a method that converts the protein (spike) sequences into the sequence similarity network (SSN). We can then use SSN as an input for the classical algorithms from the graph mining domain for the typical tasks such as classification and clustering to understand the data. We show that the proposed alignment-free method is able to outperform the current SOTA method in terms of clustering results. Similarly, we are able to achieve higher classification accuracy using well-known Node2Vec-based embedding compared to other baseline embedding approaches.
\end{abstract}


\section{Introduction}
Because of the COVID-19 pandemic, a lot of sequence data is publicly available on databases such as GISAID~\footnote{\url{https://www.gisaid.org/}}. This motivated the researchers to evaluate the data using multiple tools for the purpose of advanced analysis. Networks are one tool to this end and have found many applications in bioinformatics.
More specifically, sequence similarity networks are useful for the analysis of biological data using well-established classification and clustering methods. Clustering the sequences based on the lineages, locality, and time could help us understand the evolution and spread of different lineages, which indeed helps the biologists and relevant government authorities to take appropriate measures in advance.

\begin{figure}[h!]
  \centering
  \includegraphics[scale=0.23]{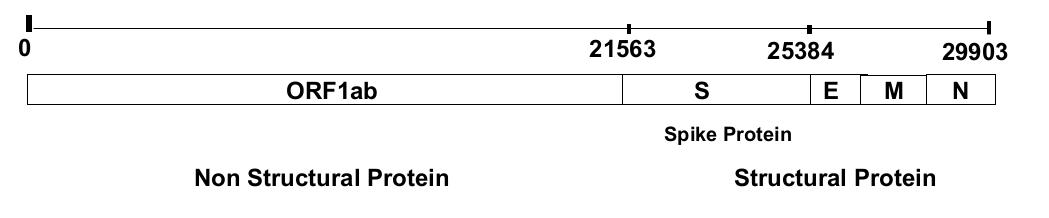}
  \caption{The full-length genome sequence contains structural and non-structural parts.  We are interested in spike (s) region as mutations related to coronavirus happen in that region.}
  \label{fig_spike_seq_example}
\end{figure}

Some effort has been made recently to classify and cluster sequences
based on hosts~\cite{kuzmin2020machine,ali2022pwm2vec} and
lineages~\cite{ali2021k,tayebi2021robust,ali2021spike2vec}.
Clustering approaches can help in identifying novel and rapidly
growing lineages, while classification can assist in keeping the track
of existing ones. Since the spike (S protein) sequence (see
Figure~\ref{fig_spike_seq_example}) of a coronavirus is the point of contact
to the host cell, it is a vital characteristic of this type of
virus~\cite{li2016structure,walls2020structure}. Therefore, the spike
the region is often the focus of the literature (rather than the entire
genome) for studying the behavior of coronaviruses, both in terms of
host and lineage
specificity~\cite{kuzmin2020machine,ali2021effective}. However, for a
spike sequence to be a compatible input to machine learning (ML)
models, it must be transformed into a fixed-length numerical
representation, known as the feature vector. There are many methods
proposed in the literature to produce such a representation of a spike
sequences, such as one-hot encoding~\cite{kuzmin2020machine}, $k$-mers
based encoding~\cite{ali2021spike2vec}, and position weight matrix
based encoding~\cite{ali2022pwm2vec}.

Our contributions to this paper are the following:
\begin{enumerate}
\item A new method to convert the biological sequences into a sequence similarity graph is proposed, which can be used to study the sequences efficiently.
\item Using the clustering algorithms, we show that the proposed model outperforms the current SOTA approach using different internal clustering evaluation metrics.
\item Using different embedding representation methods for nodes, we show that node classification using Node2Vec gives higher accuracy than other embedding approaches.
\end{enumerate}

The remaining paper is structured as follows.
Section~\ref{sec_related_work} contains related
work. Section~\ref{sec_proposed_approach} contains the proposed
approach. Section~\ref{sec_experimental_evaluation} contains
experimental setup and dataset
statistics. Section~\ref{sec_results_discussion} contains results and
discussion. Section~\ref{sec_conclusion} concludes this paper.

\section{Related Work}
\label{sec_related_work}
In the biology domain, sequence analysis using the Phylogeny-based method~\cite{Dhar2020TNet} is an important task. The method that uses substring (called mers) of length $k$ (hence $k$-mers) counts was first explored in~\cite{Blaisdell1986AMeasureOfSimilarity} for phylogenetic applications. Authors in~\cite{Blaisdell1986AMeasureOfSimilarity} proposed a  phylogenetic tree-based method from the sequences (non-coding and coding regions). In recent years, since we have had a huge availability of sequencing data due to the coronavirus pandemic, the sequence analysis problem attracts the attention of researchers~\cite{Krishnan2021PredictingVaccineHesitancy,chourasia2022reads2vec,chourasia2022clustering}. Few methods proposed for supervised tasks are alignment-free. Similarly, some approaches depend on the alignment of sequences for both supervised and unsupervised tasks.

The conversion of sequences into embeddings is an important step in the machine-learning pipeline while working with supervised and unsupervised tasks. Numerical embedding generation is important in many fields like time series forecasting~\cite{ali2019short,ali2019short_AMI}, analysis of graphs~\cite{ali2021predicting,mansoor2022impact}, electromyography~\cite{ullah2020effect}, clinical data analysis~\cite{ali2021efficient}, and network security~\cite{ali2020detecting}. 
A method to convert the aligned sequences into fixed-length embeddings using a position weight matrix is proposed in~\cite{ali2022pwm2vec}. Their method shows improvement in terms of predictive performance. However, the sequence alignment step in their pipeline makes it difficult to use for a large number of sequences. 
A method based on $k$-mers spectrum for supervised analysis of biological sequences is proposed in~\cite{ali2021spike2vec,ali2022benchmarking}. For unsupervised analysis, many recent studies proposed embedding methods for the biological sequences~\cite{ali2021effective,tayebi2021robust,ali2022efficient}.

Using kernel matrix for supervised sequence analysis is another domain explored in the literature. A kernel (gram) matrix is generated in this case that corresponds to pairwise distances between sequences~\cite{ali2022efficient, Kuksa_SequenceKernel}. This gram matrix is then used with kernel-based classifiers for supervised analysis. A kernel matrix-based method is proposed in~\cite{ali2021k} for biological sequence classification. Although the proposed method shows promising results, it is not applicable in real-world scenarios due to the expensive storage cost of the kernel matrix.

\section{Proposed Approach}\label{sec_proposed_approach}
In this section, we describe the overall approach that we are using in this paper. Our proposed method is divided into different steps described below.

\subsection{Generating Fixed-Length Numerical Representation}
\begin{figure}[h!]
  \centering
  \includegraphics[scale = 0.38]{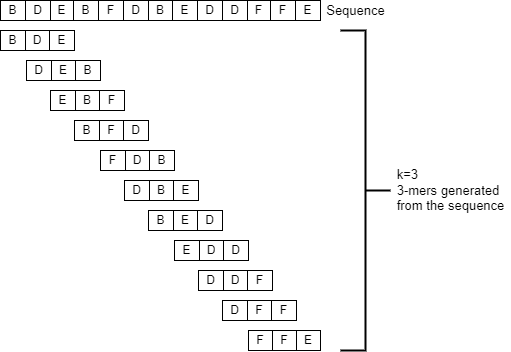}
  \caption{$k$-mers example for $k=3$.}
  \label{fig_kmer_flow}
\end{figure}

Given a set of spike sequences and attributes (coronavirus lineages) related to them, we want to generate \textit{sequence similarity network} (SSN). SSNs is a network in which nodes are sequences and
edges show the top $K$ nearest neighbors to any given sequence. Since the KNN algorithm works in euclidean space, the first step is to convert the biological sequences into a fixed-length numerical representation. For this purpose, we use a method called $k$-mers~\cite{ali2021k} (also called n-gram in the NLP domain) as given in Figure~\ref{fig_kmer_flow}. 
\begin{definition}[$k$-mers]
Given some sequence $\sigma$ on alphabet $\Sigma$, we generates
substrings (also called mers) of length $k$, \ie, $k$-mers (using sliding window approach. In
$\Sigma$, we have the following 20 characters (amino acids)
``ACDEFGHIKLMNPQRSTVWY".  
\end{definition}
For any sequence of length $N$, the
total $k$-mers are:
\begin{equation}
  \text{total $k$-mers = }(N - k) + 1
\label{eq:totalkmers}
\end{equation}

We use $k=3$ for the $k$-mers, computed using the validation set.
After generating the $k$-mers for a given sequence, we create the frequency vector (numerical representation) of length $\Sigma^k$ (i.e., length comprised of all possible $k$-mers in $\Sigma$ of length $k$), which contains the count/frequencies of $k$-mers. The pseudocode to compute the frequency vectors is given in Algorithm~\ref{algo_freq_vec}.

\begin{algorithm}[h!]
  \caption{ComputeFrequencyVector}
    \label{algo_freq_vec}
	\begin{algorithmic}[1]
	\State \textbf{Input:} Set $\mathcal{S}$ of $k$-mers on alphabet $\Sigma$, the size of mers $n$
  \State \textbf{Output:} Frequency Vector $V$
    \State combos = GenerateAllKmersCombinations($\Sigma$, n)     
    \State $V$ = [0] * $\vert \Sigma \vert^{n}$ \Comment{$ \text{Total length of (zero) vector}$}
    \For{ i $\leftarrow 1$ to $ \vert \mathcal{S} \vert$}
          \State idx = combos.index($\mathcal{S}$[i]) \Comment{$ \text{Find index of $i^{th}$ k-mer} $}
          \State $V$[idx] $\leftarrow$ $V$[idx] + 1 \Comment{$ \text{Increment bin by 1} $}
        \EndFor
        \State return($V$)
  \end{algorithmic}
\end{algorithm}

\subsection{Generating Sequence Similarity Network (SSN)}
\begin{figure}[h!]
    \centering
    \includegraphics[scale = 0.15]{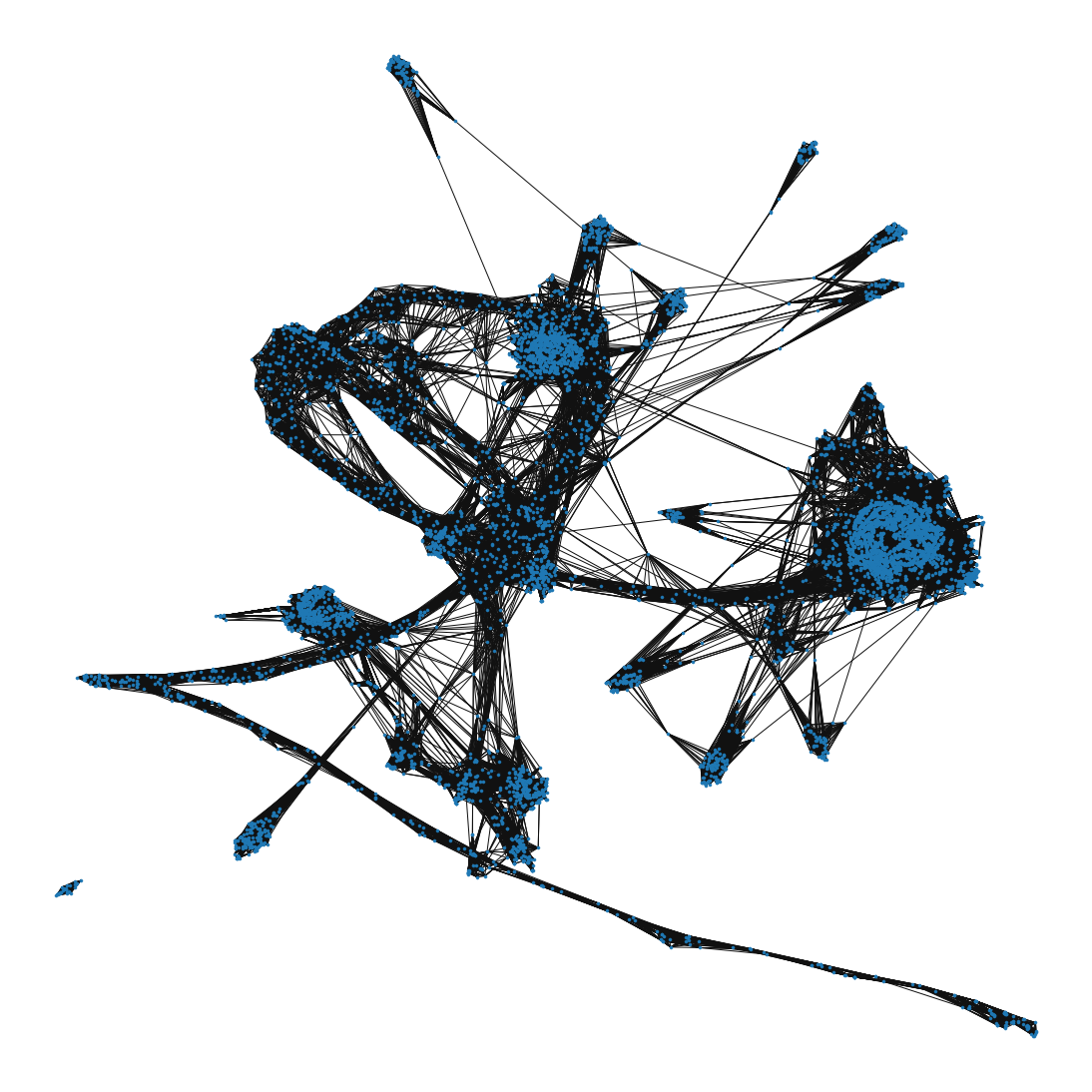}
    \caption{Input SSN Graph.}
    \label{fig_input_knn_graph}
\end{figure}

After generating the $k$-mers-based numerical representation, we use the $K$-Nearest Neighbors-based approach to generate the Sequence Similarity Network (SSN), where nodes are the sequences and edges between them show the similarity.
\begin{remark}
Note that we generate an unweighted SSN.
\end{remark}
We use $K=20$ in this case (decided using standard validation set approach~\cite{validationSetApproach}). The resultant SSN graph is given in Figure~\ref{fig_input_knn_graph}.

After generating the SSN, we use several supervised (classification) and unsupervised (clustering) methods to perform sequence analysis.

\subsection{Supervised Analysis}
Given the SSN, we perform lineage classification using different machine learning (ML) algorithms in this setting. To apply the ML algorithms on graphs, we need to get the feature embeddings for the nodes, which can be used as input to the ML classifiers. To get the numerical embeddings, we use the following methods:

\subsubsection{Locally Linear Embedding (LLE)~\cite{roweis2000nonlinear}}
Given a node $v$ from the graph along with its neighbors $N(v)$, this method assumes that $v$ is a linear combination of $N(v)$ in the embedding space (where this linear combination property must hold for all nodes in the graph). It uses the objective function to minimize the distances between the embeddings of two nodes that are neighbors to each other. The embeddings are normalized in the end to get the final representation.

\subsubsection{Laplacian Eigenmaps~\cite{belkin2003laplacian}}
Using the graph Laplacian, it generates the embedding for each node while preserving the local neighborhood information for that node. Similar to LLE, it also solves the objective function to keep the embeddings close to each other for the nodes that are neighbors. However, it involves graph Laplacian.

\subsubsection{Higher-Order Proximity preserved Embedding (HOPE)~\cite{ou2016asymmetric}}
Given the similarity matrix, it preserves the higher order proximity between the embedding of nodes by minimizing the objective function in which we minimize the squared differences between the similarity matrix and the pairs of given embeddings.

\subsubsection{Graph Factorization (GF)~\cite{ahmed2013distributed}}
Given the adjacency matrix of the graph, this method factorizes that matrix. The GF minimizes the objective function (reducing the distance between the neighboring nodes in the corresponding embeddings) to give an approximate embedding representation for a given node. Since it does not give an exact solution, the embeddings may contain some noise.

\subsubsection{DeepWalk~\cite{perozzi2014deepwalk}}
Given a pair of nodes, the goal of this method is to preserve higher-order proximity. This goal is fulfilled by maximizing the probability of observing in the random walk (previous and next top $k$ nodes).

\subsubsection{Node2Vec~\cite{grover2016node2vec}}
It is similar to DeepWalk. The difference between the two is that this approach adopts biased-random walks. This bias behavior gives a trade-off between  depth-first graph search and breadth-first graph search. Therefore it results in efficient embeddings compared to DeepWalk.

\subsection{Unsupervised Analysis}
In this setting, we perform clustering using different unsupervised clustering algorithms. We use the following clustering algorithms.

\subsubsection{MiniBatch KMeans}
It is based on the simple idea of KMeans with mini-batch
optimization. The idea of using a mini-batch is to reduce the runtime cost of the algorithm.

\subsubsection{Affinity Propagation}
Given measures of similarity between data points, it uses the concept of message passing to cluster the data. It does not require the number of clusters as input. However, it has a quadratic runtime.

\subsubsection{Mean Shift}
It is a centroid-based method that updates the candidates for centroids so that they can be the mean of the points within a given region.

\subsubsection{Spectral Clustering}
This method uses the spectrum (eigenvalues) of the kernel matrix (also called gram/similarity matrix), which is computed from the input data to perform clustering (after performing dimensionality reduction) in fewer dimensions.

\subsubsection{Ward}
It uses agglomerative hierarchical clustering along with an objective function so that merging two clusters at any iterations should satisfy the objective function (minimizing error sum of squares ESS).

\subsubsection{Agglomerative Clustering}
This method combines the pairs of clusters recursively from the data points. It uses the dendrogram-based approach.

\subsubsection{Density-Based Spatial Clustering of Applications with Noise (DBSCAN)}
Given the input data, DBSCAN groups together data points that are close to each other (nearest neighbors). It also keeps track of the outliers data points, which lies alone in low-density regions (points that do not have nearby neighbors).

\subsubsection{Ordering Points To Identify the Clustering Structure (OPTICS)}
This method searches for the core sample of high density and expands clusters from them. It is similar to DBSCAN. One of the main differences OPTICS has as compared to the DBSCAN is that it keeps cluster hierarchy for a variable neighborhood radius.

\subsubsection{BIRCH}
It works by constructing a tree data structure. In that tree, the cluster centroids behave as the leaf. Using the threshold value, clusters could be extracted from the tree. Similarly, other algorithms can be used in combination to get the final clusters, such as agglomerative clustering.

\subsubsection{Gaussian Mixture}
It contains a mixture of multiple gaussian distributions.
It is similar to the KMeans algorithm. However, it is considered a Soft clustering approach (overlapping allowed), while KMeans is considered a hard clustering algorithm (no overlapping). Given the data, this approach assigns the data points to the multivariate normal components. This assignment should maximize the component posterior probability.

\begin{remark}
Note that all clustering methods except ward and agglomerative clustering take $ k$-mers-based numerical embeddings as input. The ward and agglomerative clustering approaches take SSN as input.
\end{remark}

\section{Experimental Setup}\label{sec_experimental_evaluation}
We split the data into $70$-$30 \%$ for training and testing (held out) sets, respectively, for experimentation. We use $5$ fold cross-validation on the training data for hyperparameter tuning. The final results are computed for $30 \%$ held-out testing set. 
We use $5$ random initialization of data to avoid biases in the results. 
All experiments are performed on Windows 10 operating system, having a Core i5 processor and 32 GB RAM.  
The code is written in Python, which is available online~\footnote{\url{https://github.com/sarwanpasha/Spike2Network}}.

\subsection{Dataset Statistics}
We extracted spike sequence data from GISAID~\footnote{\url{https://www.gisaid.org/}} website (as given in~\cite{ali2022spike2signal}), which consists of $7000$ sequences of length 1274 from $22$ lineages. The proportion of lineages in the dataset is given in Table~\ref{tbl_variant_information}.

\begin{table}[ht!]
  \centering
\caption{Dataset statistics for $22$ lineages~\cite{ali2022spike2signal}. The character `-' means that information is not available.}
\resizebox{0.49\textwidth}{!}{
  \begin{tabular}{@{\extracolsep{4pt}}p{1.5cm}p{2.5cm}p{1.1cm}p{1.5cm} p{1.6cm}}
    \toprule
    \multirow{2}{*}{Lineage} & \multirow{2}{3cm}{Region of First Time Detection} & \multirow{2}{1.8cm}{Variant Name} &
    \multirow{2}{1.8cm}{No. Mut. S/Gen.} &  No. of sequences \\
    \midrule \midrule	
    B.1.1.7 & UK &  Alpha & 8/17 & \hskip.1in 3369 \\
    B.1.617.2  & India  &  Delta &  8/17  & \hskip.1in 875 \\
    AY.4   & India  & Delta  &  - & \hskip.1in 593 \\
    B.1.2   & USA  & -  & -  & \hskip.1in 333 \\
    B.1   & USA & & & \hskip.1in 292  \\
    B.1.177   & Spain &  - & -  & \hskip.1in 243 \\
    P.1   & Brazil &  Gamma &  10/21  & \hskip.1in 194  \\
    B.1.1   & UK  & & -  & \hskip.1in 163  \\
    B.1.429   &  California  & Epsilon  & 3/5  & \hskip.1in 107  \\
    B.1.526   & New York  &  Iota & 6/16 & \hskip.1in 104 \\
    AY.12   & India  & Delta  & -  & \hskip.1in 101 \\
    B.1.160   & France  & -  &  - & \hskip.1in 92 \\
    B.1.351  & South Africa  &  Beta & 9/21& \hskip.1in 81 \\
    B.1.427   & California  & Epsilon  &  3/5 & \hskip.1in 65 \\
    B.1.1.214   & Japan  & -  &  - & \hskip.1in 64 \\
    B.1.1.519   & USA  & -  & -  & \hskip.1in 56 \\
    D.2   &  Australia &  - &  - & \hskip.1in 55 \\
    B.1.221   & Netherlands & -  & -  & \hskip.1in 52 \\
    B.1.177.21   & Denmark  &  - & -  & \hskip.1in 47 \\
    B.1.258   & Germany  & -  & -  & \hskip.1in 46 \\
    B.1.243   & USA  & -  & -  & \hskip.1in 36 \\
    R.1   & Japan & -  & -  & \hskip.1in 32 \\
    \midrule
    Total & -  & -  & -  & \hskip.1in 7000 \\
    \midrule
  \end{tabular}
  }
  \label{tbl_variant_information}
\end{table}

\subsection{Elbow Method for Optimal Number of Clusters}
\begin{figure}[h!]
    \centering
    \includegraphics[scale = 0.4]{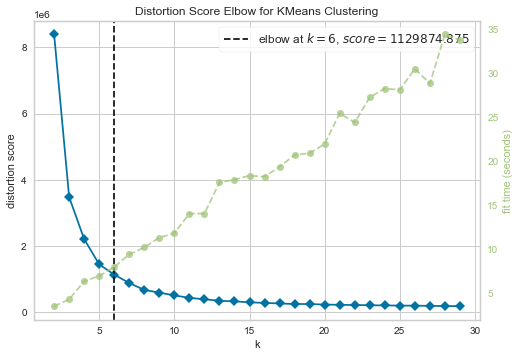}
    \caption{Elbow method using different value of $k$ to get optimal value. The notation $k$ represents the number of clusters.}
    \label{fig_elbow_method}
\end{figure}

We use a data-driven method called Elbow, to compute the optimal number of clusters~\cite{satopaa2011finding,ali2021effective}. In this method, we evaluate the trade-off between runtime and the sum of squared error and get the number of clusters where these two metrics are smaller. In this paper, we selected $4$ as the optimal cluster number (see Figure~\ref{fig_elbow_method}).

\subsection{Baseline Model}
As an unsupervised baseline, we use a method, called PWM2Vec, which is proposed in~\cite{ali2022pwm2vec}. This approach works by computing weights for each $k$-mers using the position weight matrix\cite{stormo-1982-pwm} and then concat all those weights for all $k$-mers within a sequence to get the final embedding.

\subsection{Evaluation Metrics for Clustering}
For the analysis of the unsupervised task, we use $k$-means clustering. The following evaluation metrics are used to measure the quality of clusters:

\textbf{Silhouette Coefficient}~\cite{rousseeuw1987silhouettes}: 
This metric measures the similarity of a vector with its own cluster
(cohesion) and with vectors in other clusters (separation). The value of this metric is between $-1$ (worst) to $1$ (best).

\textbf{Calinski-Harabasz Score}~\cite{calinski1974dendrite}: 
This metric works by analyzing inter-cluster dispersion and the between-cluster
dispersion (computing ratio between the two). A lower score of this metric means bad clustering and vice versa.

\textbf{Davies-Bouldin Score}~\cite{davies1979cluster}: 
This metric computes the similarity between clusters by measuring the ratio of distances within-cluster to between clusters. In this case, a lower value corresponds to better clustering and vice versa.

\subsection{Evaluation Metrics For Classification}
We use multiple classifiers to test the performance of embeddings.  The resultant feature vectors of length $200$ (using parameters tunning) are fed to
different classifiers as input.  We use Multi-Layer Perceptron (MLP), SVM, Decision Tree (DT), Naive Bayes (NB), Random Forest (RF), K-Nearest Neighbour (KNN), and Logistic
Regression (LR) algorithms.  
The evaluation metrics used to test the performance are recall, precision, accuracy, weighted F1, macro F1, and ROC area under the curve (AUC). To use the metrics designed for the binary classification task, we use the one-vs-rest approach to use them for multi-class classification.

\section{Results and Discussion}\label{sec_results_discussion}
In this section, we show the results of classification and clustering methods.
\subsection{Clustering Results}

\subsubsection{Subjective Evaluation}
\begin{figure*}[h!]
\centering
\begin{subfigure}{.33\textwidth}
  \centering
  \includegraphics[scale=0.390]{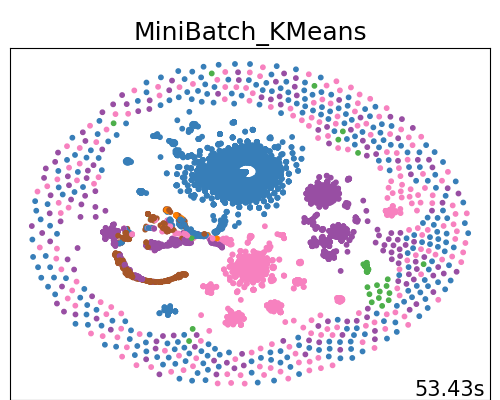}
  \caption{}
  \label{}
\end{subfigure}%
\begin{subfigure}{.33\textwidth}
  \centering
  \includegraphics[scale=0.390]{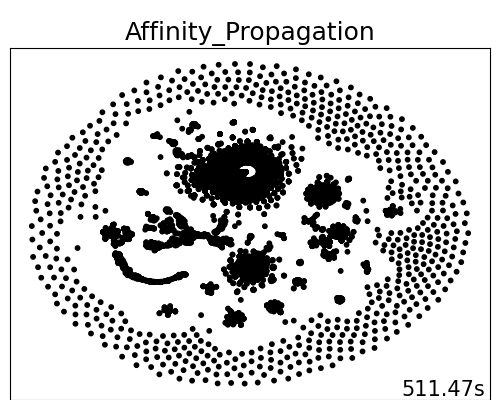}
  \caption{}
  \label{}
\end{subfigure}%
\begin{subfigure}{.33\textwidth}
  \centering
  \includegraphics[scale=0.390]{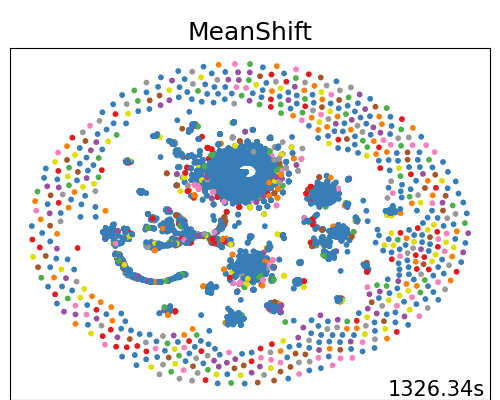}
  \caption{}
  \label{}
\end{subfigure}%
\\
\begin{subfigure}{.33\textwidth}
  \centering
  \includegraphics[scale=0.390]{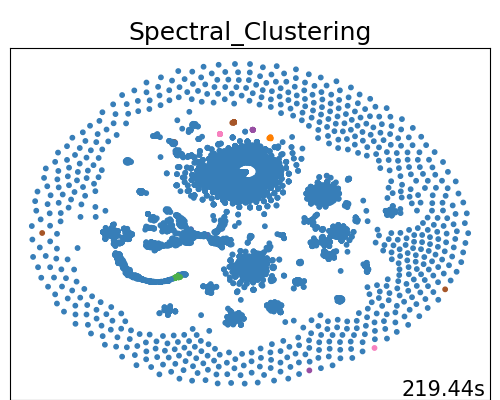}
  \caption{}
  \label{}
\end{subfigure}%
\begin{subfigure}{.33\textwidth}
  \centering
  \includegraphics[scale=0.390]{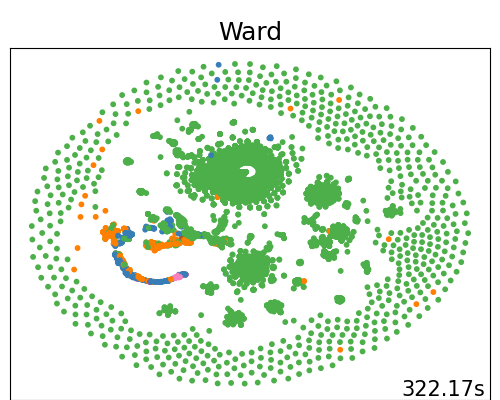}
  \caption{}
  \label{}
\end{subfigure}%
\begin{subfigure}{.33\textwidth}
  \centering
  \includegraphics[scale=0.390]{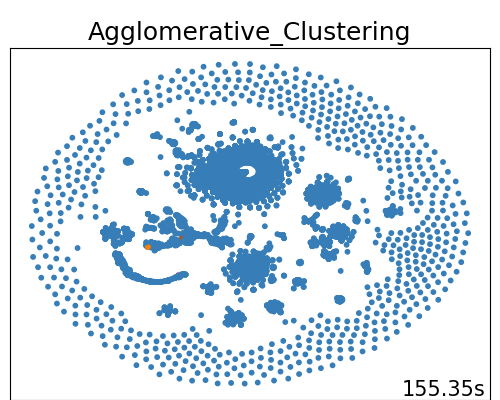}
  \caption{}
  \label{}
\end{subfigure}%
\\
\begin{subfigure}{.33\textwidth}
  \centering
  \includegraphics[scale=0.390]{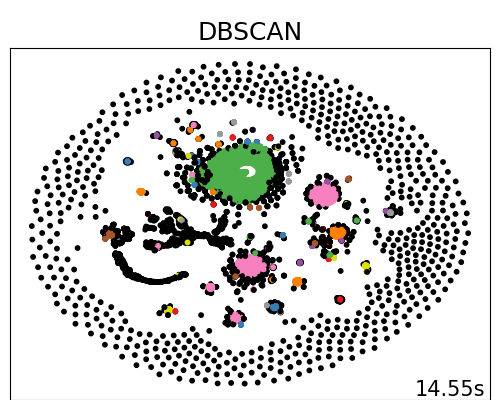}
  \caption{}
  \label{}
\end{subfigure}%
\begin{subfigure}{.33\textwidth}
  \centering
  \includegraphics[scale=0.390]{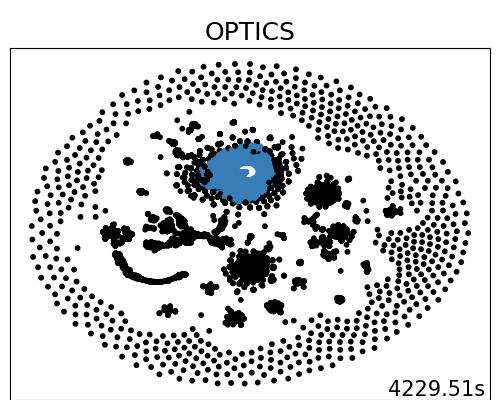}
  \caption{}
  \label{}
\end{subfigure}%
\begin{subfigure}{.33\textwidth}
  \centering
  \includegraphics[scale=0.390]{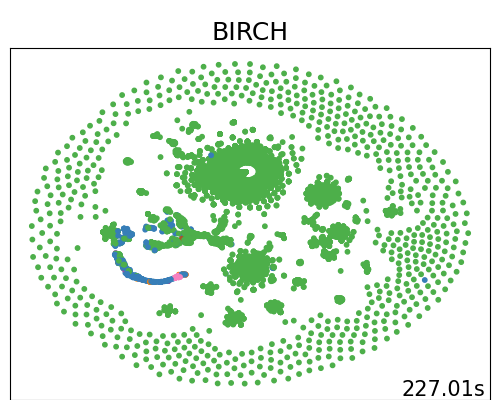}
  \caption{}
  \label{}
\end{subfigure}%
\\
\begin{subfigure}{.33\textwidth}
  \centering
  \includegraphics[scale=0.390]{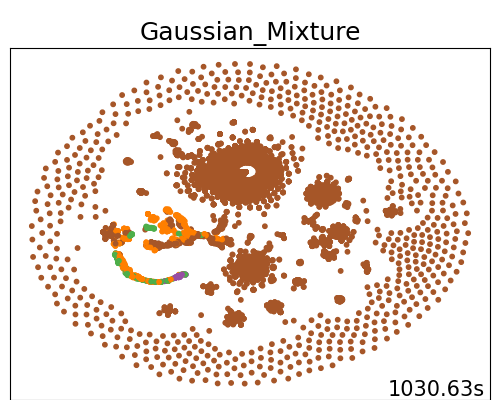}
  \caption{}
  \label{}
\end{subfigure}%
\begin{subfigure}{.33\textwidth}
  \centering
  \includegraphics[height = 3.8cm, width = 5.3cm]{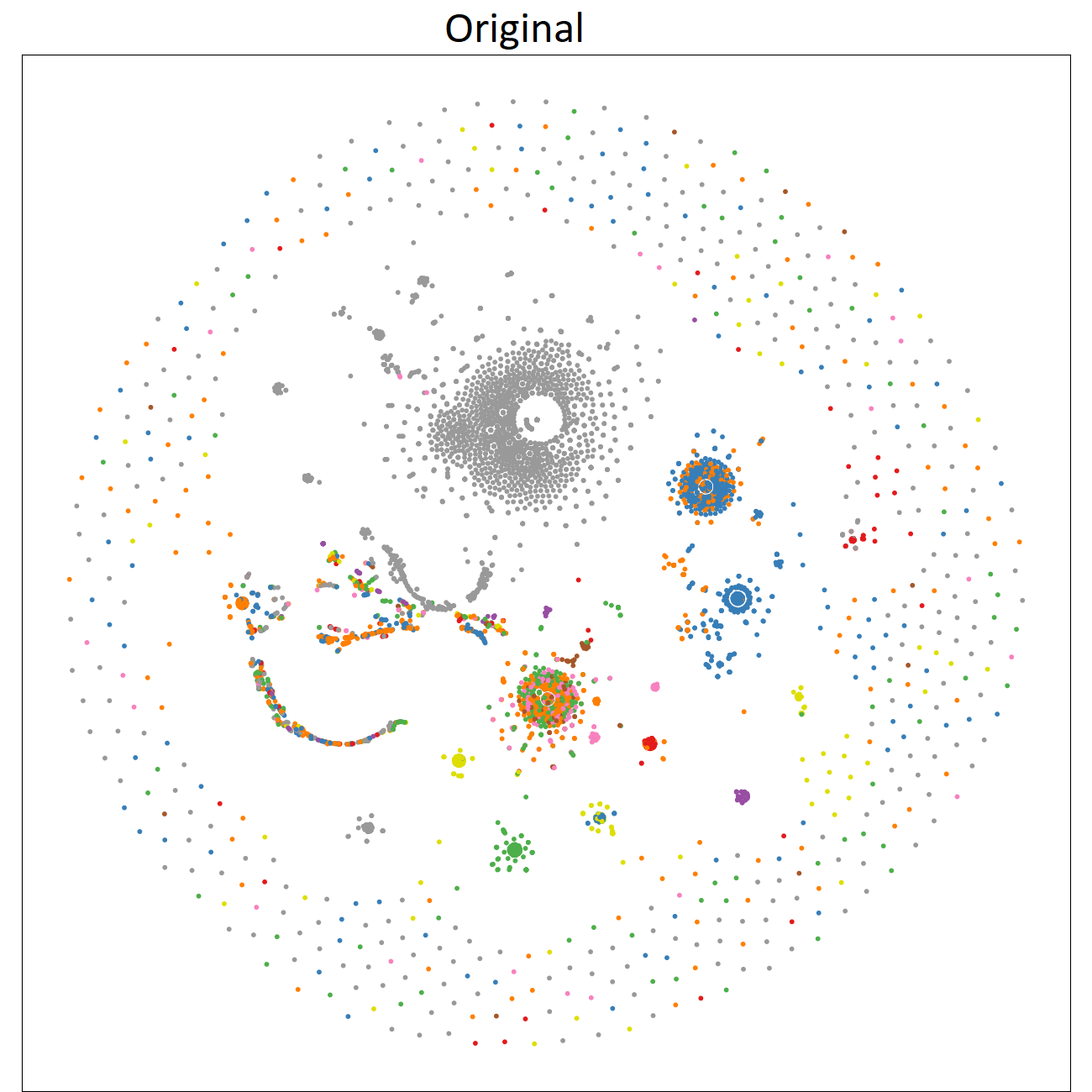}
  \caption{}
  \label{fig_org_tsne}
\end{subfigure}%
\caption{Subjective evaluation from different clustering approaches using t-SNE.}
\label{fig_overall_clustering_results}
\end{figure*}

We use t-distributed stochastic neighbor embedding (t-SNE)
\cite{van2008visualizing} to represent the data in 2-dimensions so that we can use scatter plots to visualize the data and analyze the patterns. The t-SNE method maps input data to 2D real vectors (low dimensions) while preserving the pair-wise distance between the points from high dimensions.
The subjective evaluation of different clustering methods (using t-SNE) is shown in Figure~\ref{fig_overall_clustering_results}. The t-SNE-based plot for the data with original labels is also shown in Figure~\ref{fig_org_tsne} for comparison purposes. Compared with the original plot, we can conclude that the gaussian mixture and ward give the clustering that is closely related to the original plot in some sense.

\subsubsection{Objective Evaluation}
The results for objective evaluation are given in Table~\ref{tbl_clustering_results}. In general, we can observe that there is no clear winner for the goodness of clustering. For the Silhouette coefficient, the Ward method performs the best. In the case of the Calinski-Harabasz score, the Gaussian Mixture model shows the best performance. Similarly, Agglomerative clustering shows the best performance in the case of Davies Bouldin score. PWM2Vec (the SOTA method) performs best in terms of clustering runtime because of lower dimensional embeddings.

\begin{table}[h!]
  \centering
  \caption{Objective evaluation using internal clustering quality metrics for different clustering algorithms. The best values are shown in bold.}
 \resizebox{0.49\textwidth}{!}{
  \begin{tabular}{p{1.4cm}p{1.4cm}p{1.9cm}p{1.9cm}p{1.9cm}}
    \toprule
     &  \multicolumn{3}{c}{Evaluation Metrics} & \\
    \cmidrule{2-5}
     Algorithm & Silhouette Coefficient & Calinski Harabasz Score & Davies-Bouldin Score & Clustering Runtime (Sec.) \\
    \midrule	\midrule	
    PWM2Vec & 0.477 & 1762.983 & 1.007 & \textbf{1.45} \\
    \cmidrule{2-5}
    MiniBatch KMeans & 0.133 & 1505.001 & 2.064 & 53.43 \\
    \cmidrule{2-5}
    Affinity Propagation & -0.600 & 0.111 & 1.901 & 511.47 \\
    \cmidrule{2-5}
    Mean Shift & -0.768 & 0.874 & 5.175 & 1326.34 \\
    \cmidrule{2-5}
    Spectral Clustering & -0.613 & 402.117 & 1.984 & 219.44 \\
    \cmidrule{2-5}
    Ward & \textbf{0.593} & 2657.689 & 5.152 & 322.17 \\
    \cmidrule{2-5}
    Agglomerative Clustering & 0.290 & 95.723 & \textbf{0.455} & 155.35 \\
    \cmidrule{2-5}
    DBSCAN & 0.216 & 23.174 & 1.351 & 14.55 \\
    \cmidrule{2-5}
    OPTICS & -0.196 & 285.435 & 1.548 & 4229.51 \\
    \cmidrule{2-5}
    BIRCH & 0.524 & 2242.030 & 4.757 & 227.01 \\
    \cmidrule{2-5}
    Gaussian Mixture & -0.418 & \textbf{2764.134} & 1.463 & 1030.63 \\
    \bottomrule
  \end{tabular}
  }
  \label{tbl_clustering_results}
\end{table}

\subsection{Classification Results}
The classification results for different embedding methods and classification algorithms are given in Table~\ref{tbl_classification_results}. We can observe that Node2Vec significantly outperforms all other embeddings for all evaluation metrics using SVM and KNN classifiers. This behavior shows that the embeddings generated using Node2Vec were able to preserve the structure of the network more efficiently than the other embedding approaches.

\begin{table}[h!]
    \centering
     \caption{Classification results (averaged over $5$ runs) using different evaluation metrics. The best values are shown in bold.}
    \resizebox{0.49\textwidth}{!}{
    \begin{tabular}{p{1.3cm}cp{0.7cm}p{0.7cm}p{0.7cm}p{0.7cm}p{0.7cm}p{0.7cm}|p{1.3cm}}
    \midrule
        Embed. & & Acc. & Prec. & Recall & F1 (Weig.) & F1 (Macro) & ROC AUC & Train Time (Sec.) \\
        \toprule \toprule
        
        \multirow{7}{1.5cm}{Laplacian Eigenmaps}
          & SVM & 0.47 & 0.22 & 0.47 & 0.30 & 0.03 & 0.50 & 3.45 \\
 &  NB & 0.01 & 0.11 & 0.01 & 0.01 & 0.01 & 0.50 & 0.11  \\ 
 &  MLP & 0.40 & 0.25 & 0.40 & 0.30 & 0.04 & 0.50 & 12.98  \\ 
 &  KNN & 0.37 & 0.25 & 0.37 & 0.29 & 0.04 & 0.50 & 0.33  \\
 &  RF & 0.39 & 0.25 & 0.39 & 0.29 & 0.04 & 0.50 & 7.84  \\
 &  LR & 0.47 & 0.22 & 0.47 & 0.30 & 0.03 & 0.50 & 2.18  \\
 &  DT & 0.30 & 0.27 & 0.30 & 0.28 & 0.05 & 0.50 & 1.70  \\
 
 \cmidrule{2-9}	
 
 \multirow{7}{1.5cm}{Locally Linear Embedding}
          & SVM & 0.49 & 0.24 & 0.49 & 0.32 & 0.03 & 0.50 & 3.28 \\
 & NB & 0.01 & 0.26 & 0.01 & 0.01 & 0.01 & 0.50 & 0.12 \\
 & MLP & 0.42 & 0.28 & 0.42 & 0.32 & 0.04 & 0.50 & 20.98 \\
 & KNN & 0.39 & 0.27 & 0.39 & 0.31 & 0.04 & 0.50 & 0.34 \\
 & RF & 0.36 & 0.26 & 0.36 & 0.30 & 0.04 & 0.50 & 7.64 \\
 & LR & 0.49 & 0.24 & 0.49 & 0.32 & 0.03 & 0.50 & 1.16 \\
 & DT & 0.26 & 0.27 & 0.26 & 0.27 & 0.05 & 0.50 & 1.99 \\
 
 \cmidrule{2-9}	
  \multirow{7}{1.5cm}{HOPE Embedding}
     & SVM & 0.48 & 0.23 & 0.48 & 0.31 & 0.03 & 0.50 & 2.57 \\
 & NB & 0.01 & 0.24 & 0.01 & 0.01 & 0.01 & 0.50 & 0.12 \\
 & MLP & 0.45 & 0.27 & 0.45 & 0.32 & 0.04 & 0.50 & 18.77 \\
 & KNN & 0.38 & 0.26 & 0.38 & 0.30 & 0.04 & 0.50 & 0.30 \\
 & RF & 0.37 & 0.25 & 0.37 & 0.29 & 0.03 & 0.50 & 6.68 \\
 & LR & 0.48 & 0.23 & 0.48 & 0.31 & 0.03 & 0.50 & 1.01 \\
 & DT & 0.30 & 0.26 & 0.30 & 0.28 & 0.04 & 0.50 & 1.64 \\

 \cmidrule{2-9}	
  \multirow{7}{1.5cm}{Graph Factorization}
    & SVM & 0.46 & 0.21 & 0.46 & 0.29 & 0.03 & 0.50 & 2.92 \\
 & NB & 0.37 & 0.51 & 0.37 & 0.40 & 0.31 & 0.64 & 0.11 \\
 & MLP & 0.54 & 0.51 & 0.54 & 0.53 & 0.22 & 0.60 & 18.46 \\
 & KNN & 0.69 & 0.69 & 0.69 & 0.69 & 0.51 & 0.73 & 0.39 \\
 & RF & 0.54 & 0.53 & 0.54 & 0.44 & 0.14 & 0.54 & 5.47 \\
 & LR & 0.46 & 0.21 & 0.46 & 0.29 & 0.03 & 0.50 & 1.26 \\
 & DT & 0.40 & 0.40 & 0.40 & 0.40 & 0.14 & 0.56 & 1.28 \\

 \cmidrule{2-9}	
  \multirow{7}{1.5cm}{DeepWalk}
     & SVM & 0.48 & 0.23 & 0.48 & 0.31 & 0.03 & 0.50 & 2.73 \\
 & NB & 0.46 & 0.26 & 0.46 & 0.32 & 0.03 & 0.50 & 0.13 \\
 & MLP & 0.32 & 0.26 & 0.32 & 0.29 & 0.04 & 0.50 & 23.40 \\
 & KNN & 0.38 & 0.26 & 0.38 & 0.30 & 0.04 & 0.50 & 0.33 \\
 & RF & 0.48 & 0.23 & 0.48 & 0.31 & 0.03 & 0.50 & 9.27 \\
 & LR & 0.48 & 0.23 & 0.48 & 0.31 & 0.03 & 0.50 & 1.15 \\
 & DT & 0.25 & 0.25 & 0.25 & 0.25 & 0.04 & 0.50 & 2.37 \\

 \cmidrule{2-9}	
  \multirow{7}{1.5cm}{Node2Vec}
     & SVM & \textbf{0.79} & 0.74 & \textbf{0.79} & \textbf{0.77} & 0.52 & 0.75 & 1.40 \\
 & NB & 0.63 & 0.70 & 0.63 & 0.65 & 0.48 & 0.75 & \textbf{0.10} \\
 & MLP & 0.74 & 0.74 & 0.74 & 0.74 & 0.47 & 0.73 & 19.83 \\
 & KNN & 0.76 & \textbf{0.76} & 0.76 & 0.76 & \textbf{0.55} & \textbf{0.77} & 0.32 \\
 & RF & 0.78 & 0.73 & 0.78 & 0.75 & 0.51 & 0.74 & 4.97 \\
 & LR & 0.78 & 0.72 & 0.78 & 0.75 & 0.49 & 0.73 & 3.81 \\
 & DT & 0.68 & 0.68 & 0.68 & 0.68 & 0.42 & 0.71 & 1.06 \\

        \bottomrule
    \end{tabular}
    }
    \label{tbl_classification_results}
\end{table}

To evaluate whether the computed results are statistically significant, we used the student t-test and evaluated the $p$-values for the results using the average and standard deviation results of $5$ runs.
The standard deviation results are reported in Table~\ref{tbl_std_org}. We noted that the $p$-values were $< 0.05$ in the majority of the cases and for all embedding methods (because standard deviation values are very low), hence confirming the statistical significance of the reported results. Note that we have not reported the $p$-values in this paper due to space constraints. However, we believe that they can easily be computed by anyone using the reported average and standard deviation results.

\begin{table}[h!]
  \centering
  \caption{Standard Deviation values of 5 runs for Classification results on the proposed and SOTA methods. The average results are reported in Table~\ref{tbl_classification_results}.}
    \resizebox{0.49\textwidth}{!}{
  \begin{tabular}{p{1.5cm}p{1.3cm}p{1.3cm}p{1.3cm}p{1.3cm}p{1.3cm}cp{1.3cm} | p{1.3cm}}
    \toprule
    \multirow{2}{1.1cm}{Embed. Method} & \multirow{2}{0.7cm}{ML Algo.} & \multirow{2}{*}{Acc.} & \multirow{2}{*}{Prec.} & \multirow{2}{*}{Recall} & \multirow{2}{0.9cm}{F1 weigh.} & \multirow{2}{0.9cm}{F1 Macro} & \multirow{2}{1.2cm}{ROC- AUC} & Train. runtime (sec.) \\	
    \midrule \midrule	
    
    \multirow{7}{1.5cm}{Laplacian Eigenmaps}  
     & SVM & 0.005339 & 0.005547 & 0.005339 & 0.004908 & 0.00894 & 0.003092 & 3.916014 \\
 & NB & 0.165977 & 0.047923 & 0.165977 & 0.124314 & 0.028412 & 0.021885 & 0.234008 \\
 & MLP & 0.010331 & 0.011387 & 0.010331 & 0.010000 & 0.024026 & 0.013581 & 5.913928 \\
 & KNN & 0.015266 & 0.00962 & 0.015266 & 0.014351 & 0.016323 & 0.006342 & 2.590869 \\
 & RF & 0.007615 & 0.010725 & 0.007615 & 0.008411 & 0.016425 & 0.007389 & 0.433674 \\
 & LR & 0.006048 & 0.005895 & 0.006048 & 0.006385 & 0.01425 & 0.005388 & 2.752432 \\
 & DT & 0.004937 & 0.005132 & 0.004937 & 0.004463 & 0.003758 & 0.002974 & 0.437479 \\
    \cmidrule{2-9}	
    \multirow{7}{1.5cm}{Locally Linear Embedding}  
     & SVM & 0.009732 & 0.014177 & 0.009732 & 0.011394 & 0.040983 & 0.022625 & 0.111355 \\
 & NB & 0.04711 & 0.028739 & 0.04711 & 0.034415 & 0.021785 & 0.019178 & 0.047905 \\
 & MLP & 0.004077 & 0.010426 & 0.004077 & 0.005091 & 0.020858 & 0.007274 & 2.914309 \\
 & KNN & 0.011276 & 0.010159 & 0.011276 & 0.011393 & 0.017793 & 0.007173 & 0.048912 \\
 & RF & 0.004304 & 0.01031 & 0.004304 & 0.007477 & 0.00978 & 0.007079 & 0.784592 \\
 & LR & 0.003573 & 0.00751 & 0.003573 & 0.004969 & 0.010045 & 0.005569 & 5.316077 \\
 & DT & 0.005904 & 0.00566 & 0.005904 & 0.004654 & 0.012841 & 0.005028 & 0.058772 \\
    \cmidrule{2-9}	
    \multirow{7}{1.5cm}{HOPE Embedding}  
 & SVM & 0.010442 & 0.013078 & 0.010442 & 0.011673 & 0.022631 & 0.007962 & 0.24096 \\
 & NB & 0.008815 & 0.005557 & 0.008815 & 0.00703 & 0.011761 & 0.009719 & 0.006103 \\
 & MLP & 0.007526 & 0.010713 & 0.007526 & 0.007969 & 0.020091 & 0.00948 & 2.020901 \\
 & KNN & 0.006269 & 0.006129 & 0.006269 & 0.005595 & 0.014096 & 0.00755 & 0.050194 \\
 & RF & 0.008558 & 0.010725 & 0.008558 & 0.011033 & 0.01859 & 0.006795 & 0.194081 \\
 & LR & 0.006258 & 0.013144 & 0.006258 & 0.007598 & 0.009922 & 0.00391 & 0.285232 \\
 & DT & 0.00958 & 0.011658 & 0.00958 & 0.01026 & 0.019321 & 0.007745 & 0.034749 \\
     \cmidrule{2-9}	
    \multirow{7}{1.5cm}{Graph Factorization}  
     & SVM & 0.006843 & 0.007175 & 0.006843 & 0.007533 & 0.018218 & 0.006985 & 0.127963 \\
 & NB & 0.012983 & 0.014996 & 0.012983 & 0.013997 & 0.009849 & 0.005343 & 0.002441 \\
 & MLP & 0.009597 & 0.016199 & 0.009597 & 0.008501 & 0.009785 & 0.003415 & 1.973822 \\
 & KNN & 0.012222 & 0.006539 & 0.012222 & 0.011150 & 0.012703 & 0.007168 & 0.027236 \\
 & RF & 0.004658 & 0.002715 & 0.004658 & 0.004319 & 0.014516 & 0.011376 & 0.128241 \\
 & LR & 0.005999 & 0.006079 & 0.005999 & 0.005070 & 0.010821 & 0.008120 & 0.062362 \\
 & DT & 0.002869 & 0.003766 & 0.002869 & 0.001588 & 0.017623 & 0.012498 & 0.001499 \\
     \cmidrule{2-9}	
    \multirow{7}{1.5cm}{DeepWalk}  
     & SVM & 0.014645 & 0.012340 & 0.014645 & 0.015661 & 0.024540 & 0.014596 & 1.106693 \\
 & NB & 0.041749 & 0.023858 & 0.041749 & 0.033381 & 0.025358 & 0.008237 & 0.036856 \\
 & MLP & 0.006472 & 0.008860 & 0.006472 & 0.003471 & 0.022909 & 0.010423 & 3.305528 \\
 & KNN & 0.012241 & 0.018108 & 0.012241 & 0.013407 & 0.037266 & 0.019443 & 0.089361 \\
 & RF & 0.015366 & 0.013036 & 0.015366 & 0.015552 & 0.021972 & 0.013453 & 0.217082 \\
 & LR & 0.010999 & 0.012312 & 0.010999 & 0.012642 & 0.017529 & 0.008787 & 20.67326 \\
 & DT & 0.012319 & 0.015275 & 0.012319 & 0.013563 & 0.020260 & 0.014356 & 0.346420 \\
     \cmidrule{2-9}	
    \multirow{7}{1.5cm}{Node2Vec}  
     & SVM & 0.007835 & 0.007516 & 0.007835 & 0.007297 & 0.022666 & 0.011886 & 0.215113 \\
 & NB & 0.008105 & 0.023467 & 0.008105 & 0.012723 & 0.024591 & 0.010523 & 0.039378 \\
 & MLP & 0.004398 & 0.005859 & 0.004398 & 0.005486 & 0.016899 & 0.006182 & 1.631292 \\
 & KNN & 0.013624 & 0.010276 & 0.013624 & 0.014326 & 0.01157 & 0.006457 & 0.015902 \\
 & RF & 0.004179 & 0.003099 & 0.004179 & 0.004322 & 0.010787 & 0.007188 & 0.118258 \\
 & LR & 0.009168 & 0.011003 & 0.009168 & 0.01032 & 0.000823 & 0.000216 & 0.160578 \\
 & DT & 0.008422 & 0.005511 & 0.008422 & 0.008296 & 0.017698 & 0.008793 & 0.041413 \\

    \bottomrule
  \end{tabular}
  }
  \label{tbl_std_org}
\end{table}

\section{Conclusion}\label{sec_conclusion}

In this study, we propose a method to convert the protein (spike) sequences into a graph to use well-established graph mining algorithms to analyze the biological sequences in both a supervised and unsupervised manner. We show that clustering the nodes gives better qualities of the clusters as compared to the feature engineering-based approach. For the supervised analysis, we show that Node2Vec could be used to generate the feature embeddings for the nodes, which can give better classification accuracies compared to the other embedding methods.
In the future, we will work towards testing the proposed model on more number of sequences to test the scalability. Using a neural network-based model such as GNN for the classification is another interesting future work.
We will also consider other attributes, such as
locality information and calendar values for computing richer embeddings for sequences.

\bibliographystyle{splncs04}
\bibliography{references}

\begin{thebibliography}{10}
\providecommand{\url}[1]{\texttt{#1}}
\providecommand{\urlprefix}{URL }
\providecommand{\doi}[1]{https://doi.org/#1}

\bibitem{ahmed2013distributed}
Ahmed, A., Shervashidze, N., Narayanamurthy, S., Josifovski, V., Smola, A.J.:
  Distributed large-scale natural graph factorization. In: International
  conference on World Wide Web. pp. 37--48 (2013)

\bibitem{ali2019short_AMI}
Ali, S., Mansoor, H., Khan, I., Arshad, N., Khan, M.A., Faizullah, S.:
  Short-term load forecasting using ami data. preprint, arXiv:1912.12479
  (2019)

\bibitem{ali2020detecting}
Ali, S., Alvi, M.K., Faizullah, S., Khan, M.A., Alshanqiti, A., Khan, I.:
  Detecting ddos attack on sdn due to vulnerabilities in openflow. In: 2019
  International Conference on Advances in the Emerging Computing Technologies
  (AECT). pp.~1--6 (2020)

\bibitem{ali2022pwm2vec}
Ali, S., Bello, B., Chourasia, P., Punathil, R.T., Zhou, Y., Patterson, M.:
  Pwm2vec: An efficient embedding approach for viral host specification from
  coronavirus spike sequences. Biology  \textbf{11}(3), ~418 (2022)

\bibitem{ali2019short}
Ali, S., Mansoor, H., Arshad, N., Khan, I.: Short term load forecasting using
  smart meter data. In: International Conference on Future Energy Systems. pp.
  419--421 (2019)

\bibitem{ali2022spike2signal}
Ali, S., Murad, T., Chourasia, P., Patterson, M.: Spike2signal: Classifying
  coronavirus spike sequences with deep learning. In: International Conference
  on Big Data Computing Service and Applications (BigDataService). pp. 81--88
  (2022)

\bibitem{ali2021spike2vec}
Ali, S., Patterson, M.: Spike2vec: An efficient and scalable embedding approach
  for covid-19 spike sequences. In: International Conference on Big Data (Big
  Data). pp. 1533--1540 (2021)

\bibitem{ali2022efficient}
Ali, S., Sahoo, B., Khan, M.A., Zelikovsky, A., Khan, I.U., Patterson, M.:
  Efficient approximate kernel based spike sequence classification. IEEE/ACM
  Transactions on Computational Biology and Bioinformatics  (2022)

\bibitem{ali2021k}
Ali, S., Sahoo, B., Ullah, N., Zelikovskiy, A., Patterson, M., Khan, I.: A
  k-mer based approach for sars-cov-2 variant identification. In: International
  Symposium on Bioinformatics Research and Applications. pp. 153--164 (2021)

\bibitem{ali2022benchmarking}
Ali, S., Sahoo, B., Zelikovskiy, A., Chen, P.Y., Patterson, M.: Benchmarking
  machine learning robustness in covid-19 genome sequence classification. arXiv
  preprint arXiv:2207.08898  (2022)

\bibitem{ali2021predicting}
Ali, S., Shakeel, M.H., Khan, I., Faizullah, S., Khan, M.A.: Predicting
  attributes of nodes using network structure. ACM Transactions on Intelligent
  Systems and Technology  \textbf{12}(2),  1--23 (2021)

\bibitem{ali2021effective}
Ali, S., Tamkanat-E-Ali, Khan, M.A., Khan, I., Patterson, M.: Effective and
  scalable clustering of sars-cov-2 sequences. In: International Conference on
  Big Data Research (ICBDR). pp.~1--8 (2021)

\bibitem{ali2021efficient}
Ali, S., Zhou, Y., Patterson, M.: Efficient analysis of covid-19 clinical data
  using machine learning models. Medical \& Biological Engineering \& Computing
  pp. 1--16 (2022)

\bibitem{belkin2003laplacian}
Belkin, M., Niyogi, P.: Laplacian eigenmaps for dimensionality reduction and
  data representation. Neural computation  \textbf{15}(6),  1373--1396 (2003)

\bibitem{Blaisdell1986AMeasureOfSimilarity}
Blaisdell, B.: A measure of the similarity of sets of sequences not requiring
  sequence alignment. Proceedings of the National Academy of Sciences
  \textbf{83},  5155--5159 (1986)

\bibitem{calinski1974dendrite}
Cali{\'n}ski, T., Harabasz, J.: A dendrite method for cluster analysis.
  Communications in Statistics-theory and Methods  \textbf{3}(1),  1--27 (1974)

\bibitem{chourasia2022clustering}
Chourasia, P., Ali, S., Ciccolella, S., Della~Vedova, G., Patterson, M.:
  Clustering sars-cov-2 variants from raw high-throughput sequencing reads
  data. In: International Conference on Computational Advances in Bio and
  Medical Sciences. pp. 133--148 (2022)

\bibitem{chourasia2022reads2vec}
Chourasia, P., Ali, S., Ciccolella, S., Della~Vedova, G., Patterson, M.:
  Reads2vec: Efficient embedding of raw high-throughput sequencing reads data.
  arXiv preprint arXiv:2211.08267  (2022)

\bibitem{davies1979cluster}
Davies, D.L., Bouldin, D.W.: A cluster separation measure. IEEE transactions on
  pattern analysis and machine intelligence (2),  224--227 (1979)

\bibitem{validationSetApproach}
Devijver, P., Kittler, J.: Pattern recognition: A statistical approach. In:
  London, GB: Prentice-Hall. pp. 1--448 (1982)

\bibitem{Dhar2020TNet}
Dhar, S., et~al.: Tnet: Phylogeny-based inference of disease transmission
  networks using within-host strain diversity. In: International Symposium on
  Bioinformatics Research and Applications (ISBRA). pp. 203--216 (2020)

\bibitem{grover2016node2vec}
Grover, A., Leskovec, J.: node2vec: Scalable feature learning for networks. In:
  International Conference on Knowledge Discovery \& Data Mining (KDD). pp.
  855--864 (2016)

\bibitem{Krishnan2021PredictingVaccineHesitancy}
Krishnan, G., Kamath, S., Sugumaran, V.: Predicting vaccine hesitancy and
  vaccine sentiment using topic modeling and evolutionary optimization. In:
  International Conference on Applications of Natural Language to Information
  Systems (NLDB). pp. 255--263 (2021)

\bibitem{Kuksa_SequenceKernel}
Kuksa, P., Khan, I., Pavlovic, V.: Generalized similarity kernels for efficient
  sequence classification. In: SIAM International Conference on Data Mining
  (SDM). pp. 873--882 (2012)

\bibitem{kuzmin2020machine}
Kuzmin, K., Adeniyi, A.E., DaSouza~Jr, A.K., Lim, D., Nguyen, H., Molina, N.R.,
  Xiong, L., Weber, I.T., Harrison, R.W.: Machine learning methods accurately
  predict host specificity of coronaviruses based on spike sequences alone.
  Biochemical and Biophysical Research Communications  \textbf{533}(3),
  553--558 (2020)

\bibitem{li2016structure}
Li, F.: Structure, function, and evolution of coronavirus spike proteins.
  Annual review of virology  \textbf{3},  237--261 (2016)

\bibitem{van2008visualizing}
Van~der Maaten, L., Hinton, G.: Visualizing data using t-sne. Journal of
  machine learning research  \textbf{9}(11) (2008)

\bibitem{mansoor2022impact}
Mansoor, H., Ali, S., Alam, S., Khan, M.A., Khan, I., et~al.: Impact of missing
  data imputation on the fairness and accuracy of graph node classifiers. arXiv
  preprint arXiv:2211.00783  (2022)

\bibitem{ou2016asymmetric}
Ou, M., Cui, P., Pei, J., Zhang, Z., Zhu, W.: Asymmetric transitivity
  preserving graph embedding. In: International conference on Knowledge
  discovery and data mining. pp. 1105--1114 (2016)

\bibitem{perozzi2014deepwalk}
Perozzi, B., Al-Rfou, R., Skiena, S.: Deepwalk: Online learning of social
  representations. In: International Conference on Knowledge Discovery \& Data
  Mining (KDD). pp. 701--710 (2014)

\bibitem{rousseeuw1987silhouettes}
Rousseeuw, P.J.: Silhouettes: a graphical aid to the interpretation and
  validation of cluster analysis. Journal of computational and applied
  mathematics  \textbf{20},  53--65 (1987)

\bibitem{roweis2000nonlinear}
Roweis, S., Saul, L.: Nonlinear dimensionality reduction by locally linear
  embedding. science  \textbf{290}(5500),  2323--2326 (2000)

\bibitem{satopaa2011finding}
Satopaa, V., et~al.: Finding a kneedle in a haystack: Detecting knee points in
  system behavior. In: International conference on distributed computing
  systems workshops. pp. 166--171 (2011)

\bibitem{stormo-1982-pwm}
Stormo, G.D., Schneider, T.D., Gold, L., Ehrenfeucht, A.: { Use of the
  ‘Perceptron’ algorithm to distinguish translational initiation sites in
  E. coli}. Nucleic Acids Research  \textbf{10}(9),  2997--3011 (1982).
  \doi{10.1093/nar/10.9.2997}, \url{https://doi.org/10.1093/nar/10.9.2997}

\bibitem{tayebi2021robust}
Tayebi, Z., Ali, S., Patterson, M.: Robust representation and efficient feature
  selection allows for effective clustering of sars-cov-2 variants. Algorithms
  \textbf{14}(12), ~348 (2021)

\bibitem{ullah2020effect}
Ullah, A., Ali, S., Khan, I., Khan, M.A., Faizullah, S.: Effect of analysis
  window and feature selection on classification of hand movements using emg
  signal. In: SAI Intelligent Systems Conference (IntelliSys). pp. 400--415
  (2020)

\bibitem{walls2020structure}
Walls, A.C., Park, Y.J., Tortorici, M.A., Wall, A., McGuire, A.T., Veesler, D.:
  Structure, function, and antigenicity of the sars-cov-2 spike glycoprotein.
  Cell  \textbf{181}(2),  281--292 (2020)

\end{thebibliography}

\end{document}